\documentclass[runningheads]{llncs}
\usepackage[T1]{fontenc}
\usepackage{graphicx}
\newif\ifanonymous
\anonymousfalse 

\usepackage{booktabs}
\usepackage{subcaption}
\usepackage{acro}
\usepackage{amsmath}

\begin{document}
\title{Feature extraction for LSTM-based finger-joint angle estimation from sEMG}
%
%
\ifanonymous
  \author{Anonymous Author(s)}
  \authorrunning{Anonymous Author(s)}
  \institute{Anonymous institution(s)}
\else

\author{
    Gerda de Jager\inst{1}\orcidID{0009-0009-7074-2839} \and
    Marelie H Davel\inst{1,2,3}\orcidID{0000-0003-3103-5858} \and
    Ian Thomson\inst{1}\orcidID{0000-0003-3134-8758}
}
\authorrunning{G. de Jager et al.}
%
\institute{
Faculty of Engineering, North-West University, Potchefstroom, South Africa
\and Centre for Artificial Intelligence Research, South Africa 
\and National Institute for Theoretical and Computational Sciences, South Africa
\email{38357410@mynwu.ac.za \\
\email{\{marelie.davel, ian.thomson\}@nwu.ac.za}}}
\fi

\maketitle              
\begin{abstract}
Continuous estimation of finger-joint angles from surface electromyography (sEMG) is a promising approach for controlling dexterous myoelectric prostheses.
Deep time series models have been applied successfully in this
context, typically relying on feature extraction rather than raw
signals, given limited subject-specific data.
Various time-domain and frequency-domain features can be used, 
extracted for a given window and step size.
This study systematically investigates these three design choices for Long Short-Term Memory (LSTM) regression on the Ninapro DB2 dataset. 
Window sizes from 300–1200 ms and step sizes from 10–100 ms were evaluated in combination with six time- and frequency-domain features.
Models were evaluated using regression performance metrics, inference time, and memory footprint to assess feasibility on embedded microcontroller hardware suitable for a prosthetic device.
A 300 ms window with a 100 ms step, paired with either the Mean Absolute Value (MAV) or Root Mean Squared (RMS) feature, yielded the best performance (test PCC of 0.779 ± 0.080 / 0.779 ± 0.085). 
Window size, step size, and feature choice were found to interact non-trivially: jointly tuning them can improve both accuracy and computational feasibility for real-time, embedded myoelectric control.

\end{abstract}

\section{Introduction}

Smart prosthetics are able to translate human intention into movement.
Such devices rely on electronic components and control systems for actuation, rather than being purely mechanical in nature. For a natural interactive interface and intuitive prosthetic control, an efficient Human Machine Interface (HMI) system is required~\cite{chen-2021-semg-large-scale-temporal-conv,gao-2021-multi-dof-hand}, one that reacts to human intentions quickly and accurately, without placing new physical or mental burdens on the amputee~\cite{chen-2021-semg-large-scale-temporal-conv,vujaklija-2016-new-prosthetic-dev}. A practical smart prosthesis control system should therefore have a quick response time, both to reduce perceived lag and to make the prosthesis feel more intuitive to use.

Many smart prostheses, especially those for upper-limb amputees, make use of electromyography (EMG) signals to interpret human intention. These signals are electrical in nature, generated by neuromuscular activity during muscle contraction.
EMG signals are widely used in clinical diagnosis, rehabilitation of motor disability, and biomedical applications~\cite{criswell-2010-cram}.
%
EMG sensors are broadly classified into two categories: invasive, using wires inserted beneath the skin and fat to contact muscle directly, and non-invasive, using electrodes placed on the skin's surface.
In the field of wearable electronics and prostheses, there has been minimal interest in the use of invasive EMG electrodes. Even though they have excellent spatial selectivity (i.e., demonstrate robustness against cross-talk) and deliver high-quality data, they are invasive and require a specialist to place them~\cite{catacora-2023-three-doublediff}. 

Non-invasive electrodes have gained popularity because they are more comfortable to wear and easy to use~\cite{catacora-2023-three-doublediff,guerrero-2015-analysis-doubleDiff}. Among these, direct contact electrodes, compared to gel electrodes, have no removable parts and a very low maintenance cost, making them the ideal choice~\cite{guerrero-2015-analysis-doubleDiff}. 
%
Surface electromyography (sEMG) signals, captured by non-such invasive electrodes, are widely used for HMI control as they provide abundant information about the subject’s movement intentions~\cite{gao-2021-multi-dof-hand}. The main advantages of sEMG signals over other bio-signals, such as eye movement, are ease of capture and lower data dimensionality~\cite{bi-2019-review-emg,sawangjai-2021-eeganet-ocular}. Due to these advantages, sEMG has been applied in various ways, including wheelchair control~\cite{kundu-2018-hand-wheelchair}, stroke rehabilitation~\cite{ambrosini-2021-robotic-emg-training,lee-2022-emg-stroke-training}, classification of hand movements~\cite{Hudgins-features,Ozdemir}, and estimation of hand grasping ability~\cite{li-2015-identification-finger-force,ngeo-2014-continuous-kinematics}.

There is an abundance of research on classification of gestures; however in recent years the focus has been slowly shifting towards regression of joint angles to better accommodate amputees. When it comes to prosthetic control, classification systems restrict the user to a limited set of movements (for example, three programmable motions: grasping, pinching, and wrist rotation) and lack continuous and natural gesture movement~\cite{guo-2019-lstm-joint-angles,koch-2020-regression-rnn}. In contrast, regression systems enable the smooth, synchronous movement required for daily activities by estimating joint angles in the hand and wrist and allowing the prosthetic to move accordingly~\cite{guo-2019-lstm-joint-angles,koch-2020-regression-rnn}. 

To predict these angles, sEMG signals can be seen as sequences of values, making Recurrent Neural Networks (RNNs) well suited to processing them~\cite{Goodfellow-et-al-2016-deep-learning}. 
One such model often used is the Long Short-Term Memory (LSTM) model~\cite{anam-2020-estimation,guo-2019-lstm-joint-angles,ilyas-2022-evaluation-gated-recurrent}, originally proposed by Hochreiter and Schmidhuber~\cite{hochreiter-1997-long-short-term-memory}. This model makes use of gates and internal recursion to dynamically control the time scale
~\cite{Goodfellow-et-al-2016-deep-learning}. Other architectures that have also been employed for the task of finger angle regression include Temporal Convolutional Networks (TCNs)~\cite{chen-2021-semg-large-scale-temporal-conv} and Convolutional Neural Network (CNN)-attention models~\cite{Putro-2024-Estimating-finger-joint,geng-2022-cnn}.

The input to these models is rarely the raw sEMG data, but rather extracted features~\cite{gao-2021-multi-dof-hand,Ozdemir,Putro-2024-Estimating-finger-joint}. In this context, various Time Domain (TD), Frequency Domain (FD), and joint Time-Frequency Domain (TFD) features have been investigated~\cite{Hudgins-features,Ozdemir}. Feature extraction relies on a window and a step size. The window defines a contiguous block of data points, typically specified as a time duration or number of samples, from which a feature is extracted. The step size determines how far the window moves along the time axis between extractions, with overlapping windows resulting when the step is smaller than the window. Together, the window size determines how much data is summarised into each feature, and the step size determines the rate at which features are supplied to the model; both have been shown to affect model performance
~\cite{anam-2020-estimation,avian-2022-estimating-finger-joint}. 

The window and step size also directly affect the response time of the prosthesis. The window size determines the initial latency before sufficient data has been acquired to generate the first prediction, whereas all subsequent predictions are delayed only by the step size. Farrell and Weir \cite{farrell-2007-optimal-controller-delay} found myoelectric prosthesis performance degrades noticeably when the controller delay exceeds approximately 100–125 ms. For all predictions after the first, the controller delay is the sum of the step size, feature extraction, inference, and prosthesis actuation time. This total should therefore not exceed 125 ms, the upper end of this range. We select a maximum step size of 100 ms, which leaves 25 ms for the remaining three components.
%
%
The system should have a low memory footprint, allowing a wider range of microcontroller units (MCUs) to be viable for the prosthesis implementation and reducing overall cost. This means smaller model architectures and features with a lower memory requirements are favoured in this work.


This work builds on that of Long et al.~\cite{long-2023-transfer-learning}, whose focus was on improving, via transfer learning, the accuracy of a multi-subject model when applied to an individual subject. To do this, they used a group of baseline models, including an LSTM, alongside their own LSTM aggregation module with multi-scale convolution (LSTA-conv). Their work began with single-subject models and systematically progressed towards multi-subject models.

This work adopts the LSTM architecture for several reasons. As introduced earlier, sEMG signals are naturally represented as sequences, and the LSTM's gating mechanism is well suited to modelling such temporal dependencies~\cite{Goodfellow-et-al-2016-deep-learning}. The architecture is also comparatively simple to implement and has been widely adopted for finger-joint angle regression from sEMG, including on the NinaPro dataset used in this work~\cite{guo-2019-lstm-joint-angles,anam-2020-estimation,ilyas-2022-evaluation-gated-recurrent,long-2023-transfer-learning}, providing a well-established point of comparison. Finally, LSTMs are generally considered more lightweight than more recent attention- or convolution-augmented alternatives~\cite{Putro-2024-Estimating-finger-joint,geng-2022-cnn,chen-2021-semg-large-scale-temporal-conv}, making them better suited to the low memory footprint required for embedded, real-time deployment on a prosthetic device.

Related studies employ a range of feature extraction techniques, and some report model training time~\cite{anam-2020-estimation,avian-2022-estimating-finger-joint,guo-2019-lstm-joint-angles,long-2023-transfer-learning}, a metric with little bearing on a deployed prosthesis, since training is completed prior to deployment. Model inference time, which directly affects real-time performance, is reported far less consistently.
%
Similarly,
very little information is given in regards to Hyperparameter (HP) tuning~\cite{anam-2020-estimation,avian-2022-estimating-finger-joint,guo-2019-lstm-joint-angles,long-2023-transfer-learning}. HP tuning is the process of incrementally improving the performance of a model by altering its HPs after most of the essential data cleaning, model design, and training configurations have been completed~\cite{tuningplaybookgithub}. 


We consider LSTM modelling of sEMG for finger-joint angle regression and examine the implications of different window sizes, step sizes, and features. Our contributions are as follows:

\begin{itemize}
    \item We investigate the feasibility of using various feature types and extraction parameters (window size, step size, and feature type) in an embedded systems environment with a strict controller delay constraint.
    
    \item We analyse the impact of various combinations of window sizes, step sizes, and features on model performance.
    
    \item We demonstrate the importance of a strict validation protocol and consistent HP tuning for reliable analysis.
    
\end{itemize}




\section{Methods and Approach}

This section introduces the task in more detail, and describes the experimental approach followed. The dataset used, related performance metrics, and baseline results benchmarked against are described first, followed by the overall experimental approach, including subject selection, dataset pre-processing, and HP tuning protocols.

\subsection{Dataset}
\label{sec:dataset}

The Non-Invasive Adaptive Hand Prosthetics (Ninapro) datasets~\cite{Ninapro-2025} are publicly available multimodal databases aimed at fostering machine learning research on human, robotic, and prosthetic hands. 
Where kinematic information is included, subjects are recorded wearing a CyberGlove, a sensorised glove used to capture finger-joint angles, as in Figure \ref{fig:cyber-glove-background}. The figure shows 22 angle sensors, or strain gauges, with their placement indicated by the stitching in the glove. For this work, only the first two joints of each finger are predicted, matching the joints predicted by Long et al.~\cite{long-2023-transfer-learning}.

Of the NinaPro collection of databases, DB2 is of particular interest to this study and is summarised in Table \ref{tab:db2-characteristics}. The database comprises 40 subjects, 29 male and 11 female and uses Delsys Trigno Wireless equipment, with 12 channels, a 20 to 450 Hz bandpass filter, and a 2 kHz sampling frequency at 16 bits. The exercises are labelled B, C, and D, corresponding to basic finger movements, grasping and functional movements, and force patterns. Exercises B and C contain between 12 and 23 hand gestures, where each gesture is repeated six times per subject. Consistent with Long et al.~\cite{long-2023-transfer-learning}, this work focuses on exercise C, specifically on the six gestures numbered 1, 2, 5, 7, 10, and 20, which cover a range of grasping motions.

\begin{table}[t]
    \centering

    \begin{minipage}{0.58\textwidth}
        \centering
        \caption{NinaPro DB2 characteristics~\cite{Ninapro-2025}.}
        \begin{tabular}{ p{0.55\textwidth} | p{0.45\textwidth} }
            \toprule
            Parameter & Value \\
            \midrule
            Participants (male/female) & 40 (29/11)\\
            
            Electrodes & 12 \\
            
            EMG sensor & Delsys Trigno Wireless EMG system  \\
            
            Sampling frequency & 2 kHz \\
            
            Exercises & B, C, D \\
            \bottomrule
        \end{tabular}
        \label{tab:db2-characteristics}

    \end{minipage}
    \hfill
    \begin{minipage}{0.32\textwidth}
        \centering

        \includegraphics[width=\textwidth]{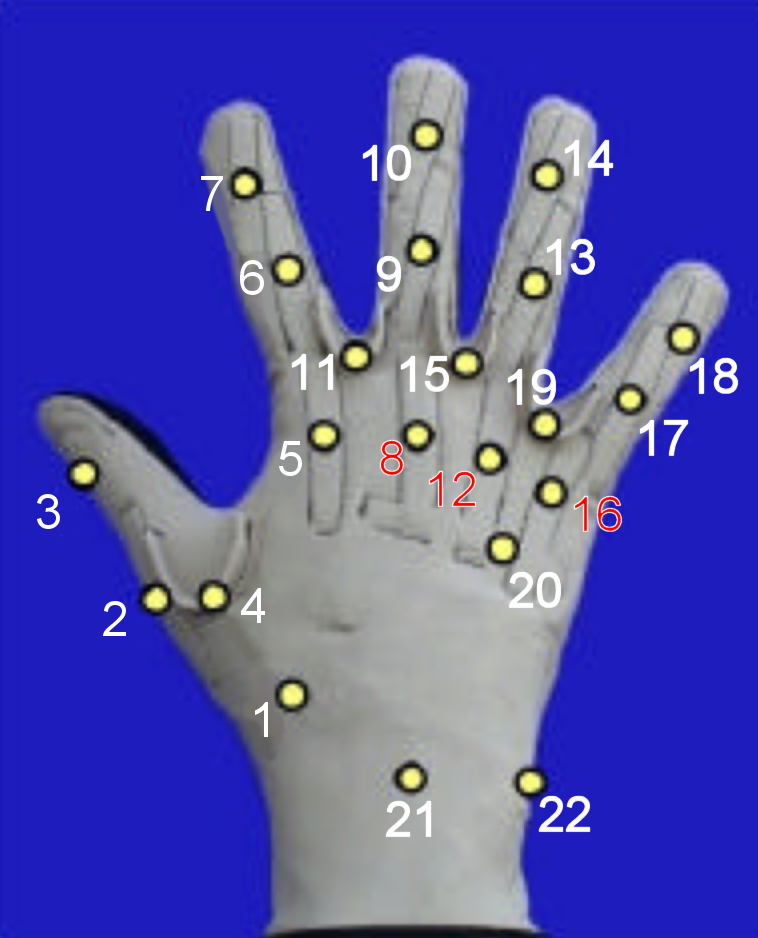}

        \captionof{figure}{NinaPro CyberGlove II with angle sensors indicated 
        \cite{Ninapro-2025}.}
        \label{fig:cyber-glove-background}

    \end{minipage}

\end{table}

\subsection{Performance Metrics}
\label{sec:performance metrics}

Three complementary regression performance metrics are used in this work: coefficient of determination ($R^2$), Pearson Correlation Coefficient (PCC), and Root Mean Square Error (RMSE).
The $R^2$ assesses the ability of a model to predict or explain the outcome in a linear regression setting, indicating the proportion of the variance in the output variable that is predicted or explained by the predictor variable \cite{britannica-coefficient-of-determination}.
The PCC measures the direction and strength of linear association between two variables. PCC values range from -1 to 1, with 1 indicating a positive linear relationship, -1 a negative linear relationship, and 0 meaning there is no linear correlation between the variables.
The RMSE measures the average magnitude of the error between predicted and true values, in the same units as the target variable.
Additional performance metric detail is provided in Appendix \ref{app:metrics}.

\subsection{Benchmarking}
\label{sec:benchmark results}

The test results reported by Long et al.~\cite{long-2023-transfer-learning} for their baseline, single-subject LSTM model, alongside their main contribution (LSTA-conv), are shown in Table \ref{tab:hpt-feature-final-stateless-rms-mav}. 
We duplicate the architecture and setup of the baseline LSTM of Long et al.~\cite{long-2023-transfer-learning} as far as possible.
Differences between our baseline and that of Long et al.~\cite{long-2023-transfer-learning} include possible differences in HP and architectural choices where specifics were not provided. In addition, data pre-processing was adapted (discussed in Section~\ref{sec:dataset pre-processing}).
The authors also used a static window size (1000 ms), step size (50 ms), and a single feature (Root Mean Squared (RMS)) -- factors that
this work will investigate. 


    
    


A final difference relates to how the dataset was partitioned, as this work uses a training, validation and test set while Long et al.~\cite{long-2023-transfer-learning} only used a training and test set. 
In both cases, datasets were partitioned based on gesture repetitions, with
Long et al.~\cite{long-2023-transfer-learning} using a 4/2 training/test split, while this work uses a 4/1/1 training/validation/test split. 
Due to the mentioned differences, the works are not directly comparable, but the results from Long et al.~\cite{long-2023-transfer-learning} remain a useful point of reference.

\subsection{Experimental Approach}
\label{sec:experimental approach}

This study consists of five distinct phases:
\begin{enumerate}

\item {\bf Benchmarking:} 
Replication of the baseline system, and review of the data pre-processing choices described in Section~\ref{sec:dataset pre-processing}.
    
\item {\bf Subject Selection:} 
A smaller set of subjects is selected for experimentation. With single-subject models, each subject requires its own HP tuning process, which quickly becomes prohibitively expensive but cannot be omitted, as results were found to be sensitive to HP tuning. The number of subjects is therefore decreased rather than reducing HP tuning. This process is described in Section \ref{sec:subject selection}.

\item {\bf Exploring Window and Step Sizes}:
For the selected subjects, five initial window sizes are investigated: 300, 550, 800, 1000, 1200 ms, and three step sizes: 10, 50, 100 ms. The smallest window size is limited to 300 ms since model predictions become increasingly noisy with smaller window sizes, seen from previous experiments. The largest window is selected as 1200 ms since a larger window reduces the number of sample points and the dataset is already limited in size. At this stage, the RMS feature is extracted from each window, matching the pre-processing method used by Long et al.~\cite{long-2023-transfer-learning}

\hspace{5mm}
The step sizes are chosen to bracket the value used by Long et al.~\cite{long-2023-transfer-learning} (50 ms), including both a smaller (10 ms) and a larger (100 ms) option. 
The context of this work is for a prosthetic control system that has a minimum actuation delay of 10 ms, therefore predictions arriving faster than 10 ms would arrive before the system has time to act on them, leaving some predictions unused. Thus smaller step sizes are not considered.
The largest step size is selected as 100 ms to respect the controller delay constraint of the prosthesis. Since the prosthesis has a maximum acceptable controller delay of 125 ms, and 100 ms is allocated to data acquisition, this leaves 25 ms for pre-processing, inference, and actuation.

\item {\bf Feature Analysis:}
In this phase, six features are investigated, three from the Time Domain (TD) and three from the Frequency Domain (FD): RMS, Mean Absolute Value (MAV), Variance (VAR), Mean Frequency (MNF), Median Frequency (MDF), and Mean Power (MNP). Their equations are included in Appendix~\ref{app:features}. 
A substantially larger set of sEMG features has been documented in the literature \cite{sid-el-moctar-2024-comprehensive-review-semg}.
The six features used here were selected from this broader set for their relative simplicity of computation and their established use in related sEMG-based finger-joint angle regression work~\cite{anam-2020-estimation,avian-2022-estimating-finger-joint,long-2023-transfer-learning,Putro-2024-Estimating-finger-joint}. 
They are first tested for suitability for deployment on an embedded system, based on their computational and memory complexity. Then, the top three window-step size combinations are used in conjunction with the suitable features, to determine the best three-way combination. 

\item {\bf Extended Evaluation:} 
In the final phase, six more subjects are added to the existing six (as described in Section \ref{sec:subject selection}) and the two best-performing (based on validation metrics) features, with their respective window-step combinations, are tuned using a more extensive HP tuning protocol (protocol B as described in Section~\ref{sec:hp tuning}) to confirm results.

\end{enumerate}

\subsection{Dataset Pre-processing}
\label{sec:dataset pre-processing}

Atzori et al.~\cite{atzori-2014-electromyography-db2} performed three pre-processing steps on the Ninapro DB2 that are included in the publicly available dataset. Firstly, synchronising: the different streams (sEMG, angle, accelerometer etc.) were upsampled to 2 kHz using linear interpolation for real-valued streams or nearest-neighbours for discrete valued streams. Secondly, relabelling: by realigning the stimuli data with the streams to account for human reaction time and experimental conditions. Thirdly, filtering: the sEMG data was filtered to remove 50 Hz (and harmonics) Power Line Interference (PLI), 
using a Hampel filter.

The sEMG data is further processed in three additional steps based on the pre-processing procedure described by Long et al.~\cite{long-2023-transfer-learning}. First, it is filtered using a fourth-order, 5–450 Hz bandpass filter with a 50 Hz notch filter, to remove noise and PLI (in addition to the processing described earlier). 
The bandpass filtering is partially redundant, as the Delsys hardware already applies a bandpass filter with a 20 to 450 Hz range.
Second, the finger-angle data is smoothed using a second-order low-pass filter with a cut-off frequency of 1 Hz. 
Third, the extracted features are normalised. 
In contrast to Long et al.~\cite{long-2023-transfer-learning}, $\mu$-law scaling is not applied and min-max normalisation is replaced by z-normalisation, as both these factors were shown to improve results, in a separate set of experiments. For reference, results are included in Appendix~\ref{app:normalisation}. 

\subsection{HP Tuning}
\label{sec:hp tuning}

The HP tuning process requires three datasets: a training set to fit the parameters, a validation set to measure the impact of selected HPs, and a test set to verify the performance of the best-tuned model \cite{tuningplaybookgithub}. 
Ultimately, the goal of HP tuning is to reduce the expected error on unseen data.

Table \ref{tab:hp-tuning-protocols-a-b} describes the two HP tuning protocols used in this work. The left side of the table outlines the first tuning protocol, Protocol A, a computationally light method used in Phases 3 and 4, to maximise model performance while limiting tuning time. Protocol A tunes the learning rate, dropout and model width over 30 sweeps, applied separately to each subject.
The right side of the table outlines the second tuning protocol, B, applied in Phases 1, 2 and 5 to determine a baseline model performance, confirm results and establish the best possible performance for the chosen window-step-feature configuration. Unlike Protocol A, Protocol B also tunes weight decay and batch size, over 60 sweeps rather than 30.
{
\setlength{\tabcolsep}{6pt}
\begin{table}[t]
    \centering
    \caption{HPs and corresponding search spaces for HP tuning Protocols A and B.}
    
    \resizebox{\textwidth}{!}{
    
    \label{tab:hp-tuning-protocols-a-b}
    \begin{tabular}{l|ll|ll}
        \toprule
        & \multicolumn{2}{c|}{Protocol A} 
        & \multicolumn{2}{c}{Protocol B} \\
        \cmidrule(lr){2-5} 
        Parameter & Search Space & Distribution 
        & Search Space & Distribution \\
        \midrule

        Dropout 
        & $[0.0, 0.1, 0.2, 0.3]$ & --
        & $[0.0, 0.1, 0.2, 0.3]$ & -- \\

        Learning rate 
        & $\sigma=1,\ \mu=\log(10^{-3})$ & Log-normal
        & $\sigma=1,\ \mu=\log(10^{-3})$ & Log-normal \\

        Width 
        & $[64, 256]$ & --
        & $[64, 256]$ & -- \\

        Weight decay
        & -- & --
        & $\sigma=1,\ \mu=\log(10^{-5})$ & Log-normal \\

        Batch size
        & -- & --
        & $[4, 64, 1024]$ & -- \\

        \bottomrule

    \end{tabular}
    }
\end{table}
}
Table \ref{tab:lstm-model-hps} lists the static HPs used for the LSTM model under Protocol A; all other HPs are held constant, except the extracted feature (variable in Phase 4) and weight decay and batch size (tunable in Phases 1, 2, and 5). The input chunk [-6, 0] includes six past samples plus the current sample; the output chunk [0, 0] predicts only the current timestep.

\begin{table}[t]
    \centering
    \caption{Static HPs for the LSTM model under Protocol A. Extracted feature, weight decay, and batch size vary under Protocol B (see Table \ref{tab:hp-tuning-protocols-a-b}).}

    \resizebox{\textwidth}{!}{
    
    \begin{tabular}{p{0.25\linewidth} p{0.3\linewidth} | p{0.25\linewidth} p{0.3\linewidth}}
    
    \toprule
    Parameter & Value & Parameter & Value \\
    \midrule

    Stacked models & 5 &
    Task  & Regression \\
    
    Number of inputs & 12 channels & 
    Stateful-ness & Stateless \\
    
    Number of outputs & 10 finger angles &
    Loss function & MSE \\
    
    Maximum epochs & 200 &
    Extracted feature & RMS \\
    
    Batch size & 64 &
    Optimiser & Adam \\
    
    Weight decay & $3 \times 10^{-6}$ &
    Layernorm & True \\
    
    Input chunk & [-6, 0] &
    Residual connections & True \\
    
    Output chunk & [0, 0] & 
    Learning rate & Reduce on plateau, delta: \\
    && scheduler & 0.5, patience: 10, threshold: 20 \\
    
    \bottomrule
    \end{tabular}

    }    
    \label{tab:lstm-model-hps}
\end{table}

\subsection{Implementation}
\label{sec:implementation}

All experiments were implemented using the KnowIt codebase~\cite{theunissen2025knowitrepo}, which is designed for building and interpreting artificial neural networks that perform time series modelling. Hyperparameter sweeps, as described in Section~\ref{sec:hp tuning}, were managed and tracked using Weights \& Biases~\cite{biewald-2020-wandb}.

\section{Results}

Results are presented per phase, for each of the five phases described in Section \ref{sec:experimental approach}.

\subsection{Benchmarking}
\label{sec:benchmarking}

This subsection has two aims: to replicate the baseline LSTM of Long et al.~\cite{long-2023-transfer-learning}, estimating unclear HPs given the differences outlined in Section \ref{sec:benchmark results}; and to determine the best achievable LSTM performance on NinaPro DB2~\cite{Ninapro-2025}, through systematic pre-processing changes and Protocol B tuning across all 35 subjects. 
Table \ref{tab:hpt-feature-final-stateless-rms-mav} records validation results without HP tuning (\textit{Phase 1 Replication}) and with adjusted pre-processing plus Protocol B tuning (\textit{Phase 1 Baseline}). Replication matched Long et al.'s~\cite{long-2023-transfer-learning} results to a satisfactory, if imprecise, degree, given the differences between the two works; adjusted pre-processing and Protocol B tuning surpassed Long et al.~\cite{long-2023-transfer-learning} on $R^2$ and PCC, though not on RMSE.

\subsection{Subject Selection}
\label{sec:subject selection}

Of the 40 subjects in DB2, 35 were used for benchmarking, matching Long et al.'s~\cite{long-2023-transfer-learning} criterion of including only right-handed subjects. Due to the significant HP tuning time required per subject (Section \ref{sec:benchmarking}), this group is reduced to six for subsequent experiments: subjects are ranked from worst to best by validation loss, and two poorly-performing (6, 40), two average-performing (20, 39), and two well-performing (9, 17) subjects are selected, ensuring the group is broadly representative of the full 35-subject dataset. Table \ref{tab:hpt-feature-final-stateless-rms-mav} reports the average validation performance of these six subjects using adjusted pre-processing and HP tuning (\textit{Phase 2 Baseline}); performance falls between the other two conditions, better than the unmodified replication, but worse than with Protocol B, nearly matching the results of Long et al.~\cite{long-2023-transfer-learning}. 

\subsection{Exploring Window and Step Sizes}

\subsubsection{Performance:}

\begin{figure}[t]
    \centering
    \begin{subfigure}{0.49\linewidth}
        \includegraphics[width=\linewidth]{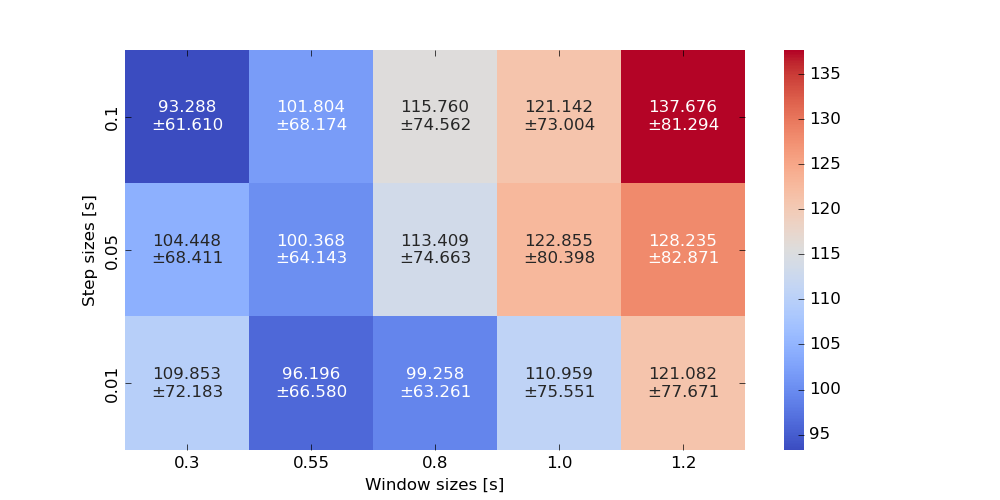}
        \caption{Loss (lower is better).}
    \end{subfigure}
    \hfill
    \begin{subfigure}{0.49\linewidth}
        \includegraphics[width=\linewidth]{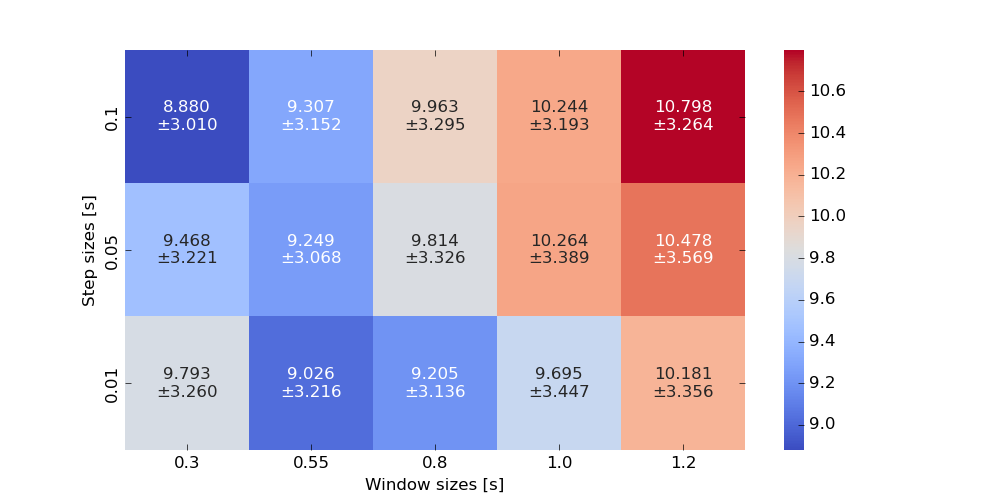}
        \caption{RMSE (lower is better).}
    \end{subfigure}
    
    \vspace{0.5em}

    \begin{subfigure}{0.49\linewidth}
        \includegraphics[width=\linewidth]{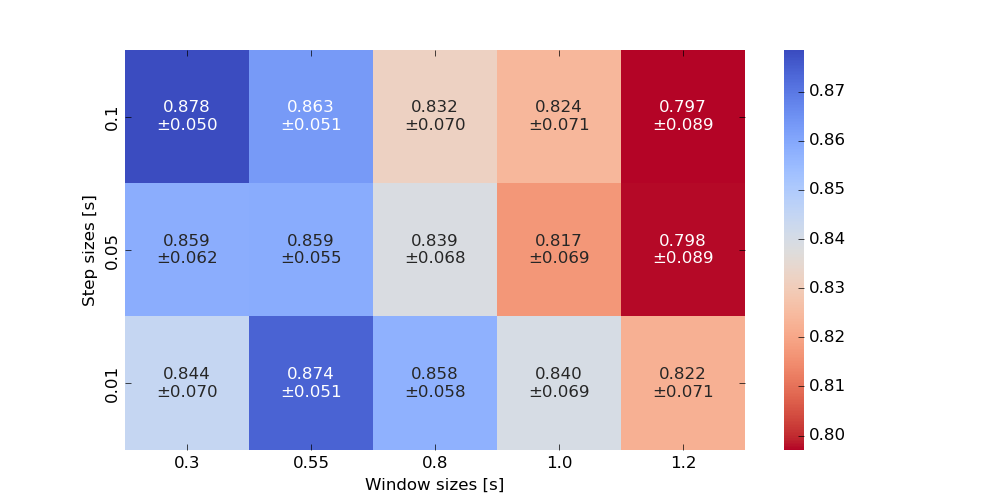}
        \caption{PCC (higher is better).}
    \end{subfigure}
    \hfill
    \begin{subfigure}{0.49\linewidth}
        \includegraphics[width=\linewidth]{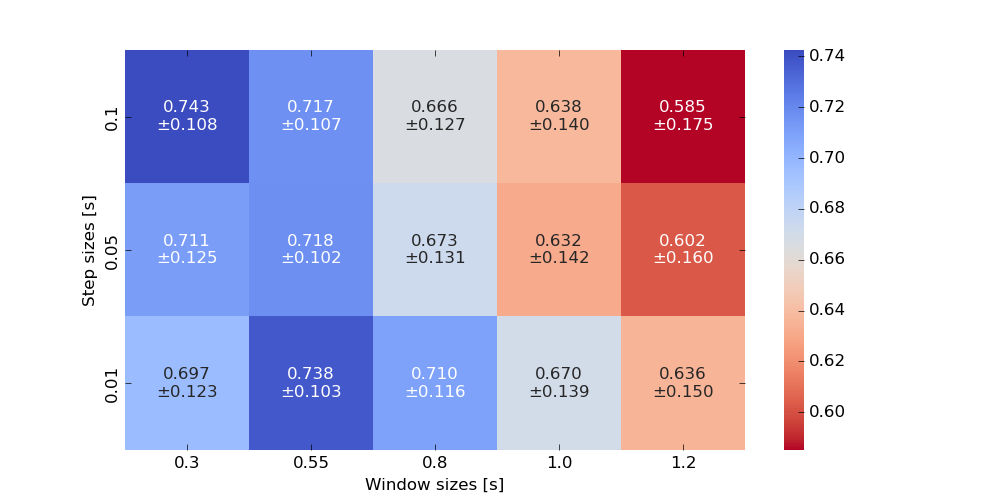}
        \caption{$R^2$ (higher is better).}
    \end{subfigure}

    \caption{Performance of window and step size combinations, using the RMS feature, evaluated on the validation set. 
    For all graphs, darker blue indicates better performance and darker red indicates worse performance.}
    \label{fig:heatmaps-window-step-exp}
\end{figure}

Figure \ref{fig:heatmaps-window-step-exp} shows the validation performance across all 15 window-step combinations, with dark blue indicating better performance; each block also shows the mean and standard deviation. Smaller window and step sizes generally perform best. 
While the 300 ms window seem to have different behaviour (best performance at 100 ms step size) this difference falls within the standard error interval. 
Based on all of the metrics, the top three combinations are 0.3-0.1, 0.55-0.01, and 0.8-0.01. 

Each of the four metrics identifies the same ranking of window-step combinations as best- and worst-performing, which is expected given that $R^2$, PCC, and RMSE each capture overlapping aspects of fit quality on the same predictions, $R^2$ and PCC both reflect how well the model's predictions track the structure of the target, while RMSE captures the magnitude of residual error. Although the four metrics agree on relative ranking, they are not redundant: PCC confirms the model correctly follows the directional trend of the target even where its absolute scale is off, $R^2$ confirms the predictions are neither flat (underfitting) nor erratic (overfitting) relative to the target's variance, and RMSE confirms the magnitude of the remaining error is small in absolute terms. The consistent agreement across all four metrics indicates that the identification of the best-performing window-step combinations are not due to an artefact of one metric's particular sensitivity.


\subsubsection{Assessment of Subject Selection:}

\begin{figure}[t]
    \centering

    \begin{minipage}{0.49\textwidth}
        \centering        
        \includegraphics[width=\textwidth]{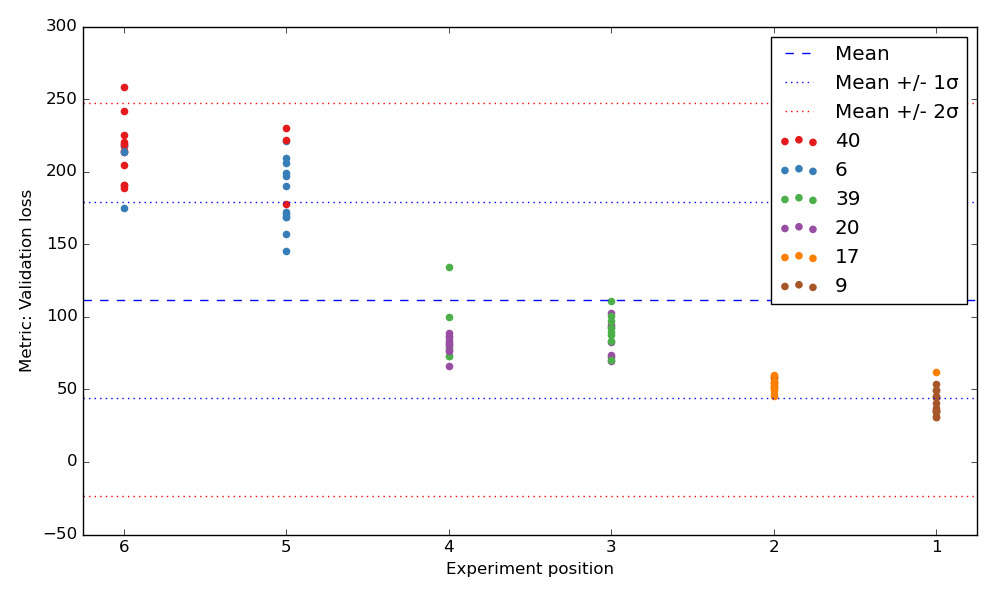}

        \subcaption{Phase 3, 15 experiments.}
        \label{fig:series-experiment-summary-window-step}

    \end{minipage}
    \hfill
    \begin{minipage}{0.49\textwidth}
        \centering
        \includegraphics[width=\textwidth]{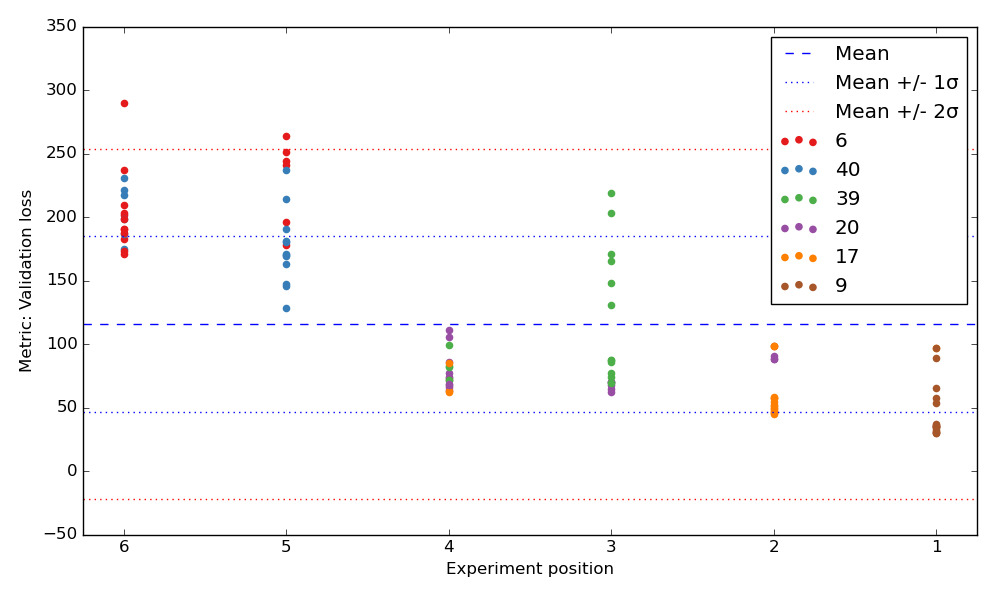}

        \subcaption{Phase 4, 18 experiments.}
        \label{fig:series-exp-summary-feature-window-step}

    \end{minipage}
    
    \caption{Subject-specific performance per experiment, with the average and standard deviation of validation loss calculated across all experiments. The x-axis ranks the subjects in order of performance for each experiment, with the worst-performing subject plotted on the left (position 6) and the best-performing subject to the right (position 1).}
\end{figure}

To assess whether the subject selection in Section \ref{sec:subject selection} was representative, Figure \ref{fig:series-experiment-summary-window-step} summarises the performance of each subject over all 15 experiments. The graph's y-axis indicates the validation loss, while the x-axis groups datasets by rank: for each experiment, the six subjects' datasets are ordered from worst to best performing (based on validation loss) and assigned to positions 1 (best) through 6 (worst). A given subject's rank position may therefore vary from one experiment to another.

Each subject is indicated by a distinct colour; ideally, the two poorest-performing subjects (40, 6) would only appear in the leftmost positions (5, 6), and the two best-performing (9, 17) only in the rightmost (1, 2), matching their designations in Section \ref{sec:subject selection}. In the figure, each pair remains within its designated group, suggesting that window and step size affect all performance groups similarly and confirming the six subjects as representative of the broader 35-subject group.



\subsubsection{Latency:}

\begin{figure}[t]
    \centering
    \begin{minipage}{0.59\textwidth}
        
        \includegraphics[width=\textwidth]{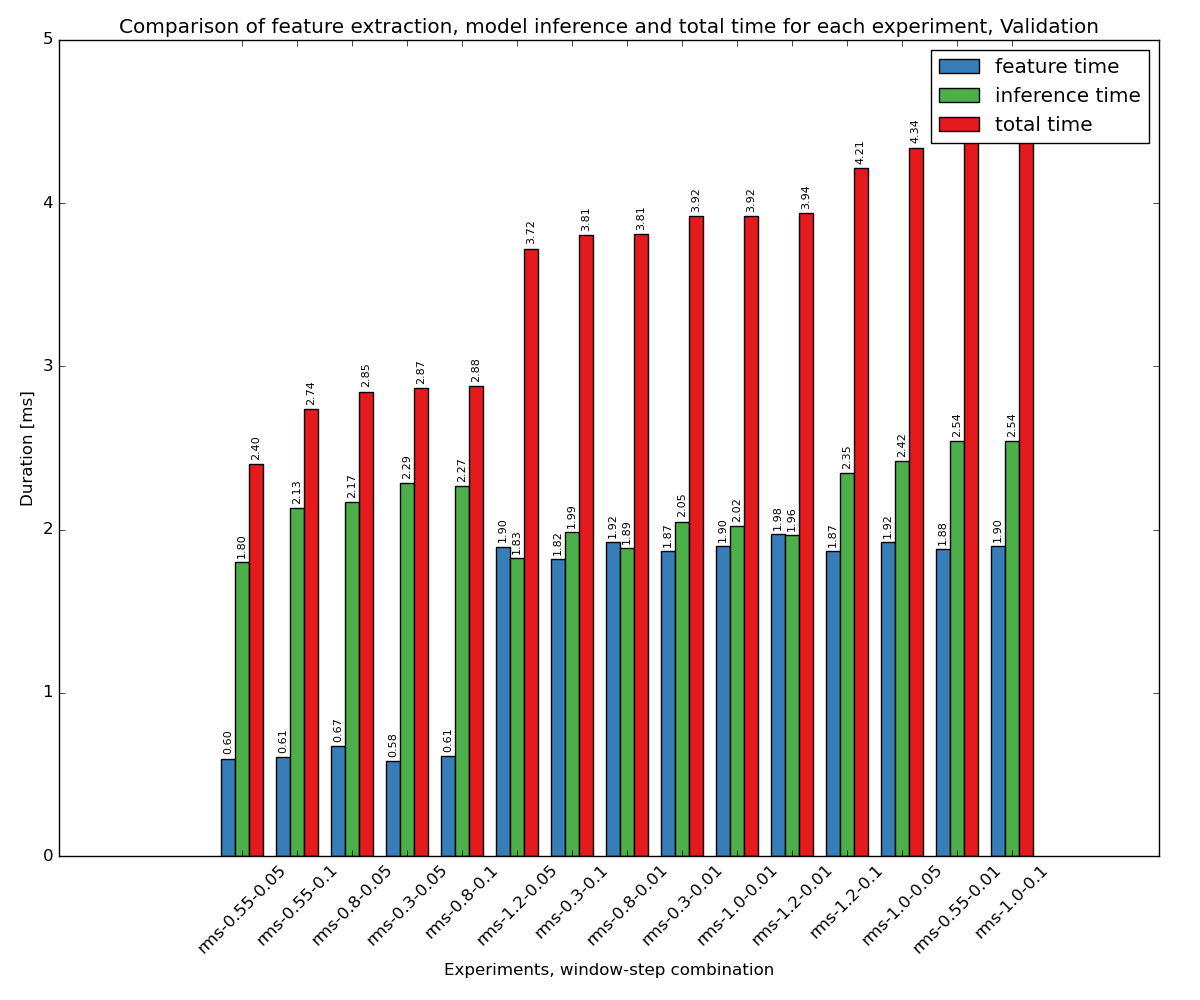}
        \caption{Timing information for each window-step combination, including feature extraction time, model inference time, and the sum of the two as the total time.}
        \label{fig:timing-bar-graph}

    \end{minipage}
    \hfill
    \begin{minipage}{0.38\textwidth}

        \includegraphics[width=\textwidth]{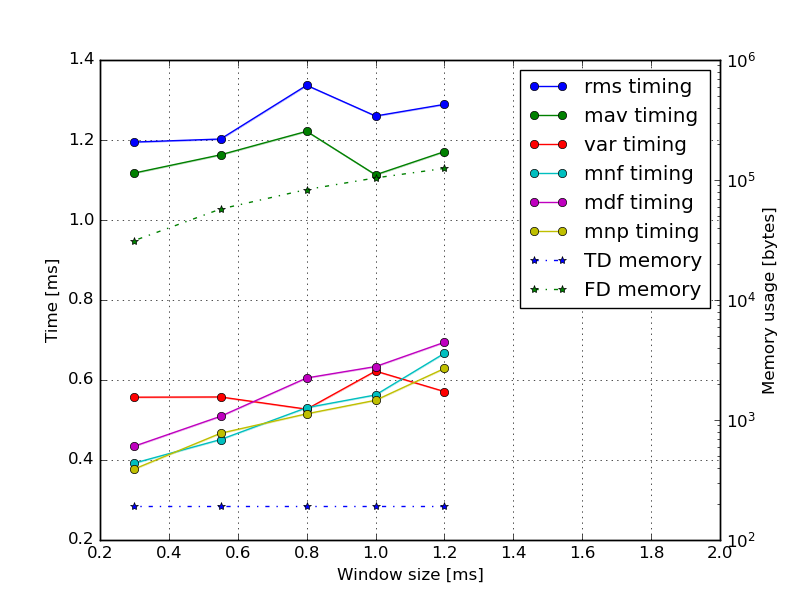}
        \caption{(Left axis) Timing graph for each feature and window size, indicated by the solid lines. (Right axis) memory usage for time and frequency domain features over various window sizes, indicated by the dashed lines.}
        \label{fig:timing-memory-features}

    \end{minipage}
\end{figure}

Figure \ref{fig:timing-bar-graph} summarises timing metrics for each window-step combination, including RMS feature extraction time, inference time, and total time. Five combinations, 0.55-0.05, 0.55-0.1, 0.8-0.05, 0.3-0.05, and 0.8-0.1, extract the feature noticeably faster than the rest, in under 0.7 ms; the remaining combinations take 1.8–1.9 ms. Inference times mostly fall between 1.8 and 2.5 ms. The quickest total time is 2.4 ms (0.55-0.05), and the slowest is 4.44 ms (1.0-0.05), both comfortably within the 125 ms controller delay constraint (even at a step size of 100 ms), with the worst case contributing at most 5 ms ($\approx$4\% of the total delay constraint).



\subsection{Feature Analysis}

\subsubsection{Computational Complexity:}
\label{sec:computational complexity}

To determine which features from the time and frequency domains are viable options for an embedded system on a smart prosthesis, the theoretical and practical computational and memory complexity of each feature is analysed. Table \ref{tab:theoretical-complexity-of-features} lists each feature's theoretical computational and memory complexities, expressed using Big-O notation (based on how operations and memory scale with the window size \textit{N}) and standard equations, given in Appendix \ref{app:features}. Features with computational or memory complexity beyond O(N) are considered high-risk for real-time deployment. 

\begin{table}[b]
    \centering
    \begin{minipage}{0.48\textwidth}

        {
        \setlength{\tabcolsep}{6pt}  
        
        \captionof{table}{Theoretical Big-O complexity characteristics for six time- and frequency-domain features.}
        \label{tab:theoretical-complexity-of-features}
        \begin{tabular}{l|l|l}
        
        \toprule
        Feature & Computational & Memory \\
        \midrule
        RMS & O(N) & O(1) \\
        MAV & O(N) & O(1) \\
        VAR & O(N) & O(1)\\
        MNF & O(Nlog(N)) & O(N)\\
        MDF & O(Nlog(N)) & O(N)\\
        MNP & O(Nlog(N)) & O(N) \\
        \bottomrule
        \end{tabular}

        }

    \end{minipage}
    \hfill
    \begin{minipage}{0.48\textwidth}

        {
        \setlength{\tabcolsep}{6pt}  
        
        \centering
        \captionof{table}{The average time taken to compute the power spectrum, required for frequency-domain features (MNF, MDF, MNP), using three different algorithms over a 1200 ms window.}
        \begin{tabular}{l|l}
        \toprule
        Algorithm & Time [ms] \\
        \midrule
        Fast Fourier Transform & 0.690 $\pm$ 0.163 \\
        Periodogram & 0.580 $\pm$ 0.136 \\
        Welch & 0.760 $\pm$ 0.172 \\
        \bottomrule
        \end{tabular}
        \label{tab:power-spectrum-timings}

        }

    \end{minipage}
\end{table}

The power spectrum was calculated using three different algorithms (Fast Fourier Transform (FFT), Periodogram, and Welch), over 10 windows of 1200 ms (2400 samples); the results are shown in Table \ref{tab:power-spectrum-timings}. 
On average the periodogram calculated the spectrum the fastest and achieved the lowest standard deviation. Thus \texttt{signal.periodogram} will be used to calculate the power spectrum for the frequency features.

Figure \ref{fig:timing-memory-features} shows the average time (left axis) versus window size for each feature, averaged over ten trials. As predicted by the theoretical analysis (Table \ref{tab:theoretical-complexity-of-features}), all features show a general increase in calculation time as window size increases. RMS and MAV take the most time to calculate, a little less than 1.4 ms at the worst. The rest of the features are calculated in less than a millisecond across all window sizes. The right-hand y-axis shows the memory usage of the FD and TD features on a logarithmic scale. The FD features require a maximum of 124,904 bytes (125 kB) at a window size of 1.2 s, while the TD remain constant at 192 bytes for every window size.

In terms of computational feasibility, none of the features required enough compute, theoretically or practically, to be disqualified from use on an MCU. Regarding memory, the largest requirement is 125 kB, and most MCUs today have 128 kB or more of Static Random Access Memory (SRAM), so it would not be difficult to find an MCU with sufficient memory for the FD calculations, model inference, and additional control systems.

\subsubsection{Performance:}

Only the three best-performing (according to validation metrics) window-step combinations (0.3-0.1, 0.55-0.01, 0.8-0.01), and the six features confirmed feasible in Section \ref{sec:computational complexity}, are used in this investigation. Testing all window-step combinations against every feature was not computationally feasible, so only the top three combinations are carried forward. 

\begin{figure}[t]
    \centering
    \begin{subfigure}{0.49\linewidth}
        \includegraphics[width=\linewidth]{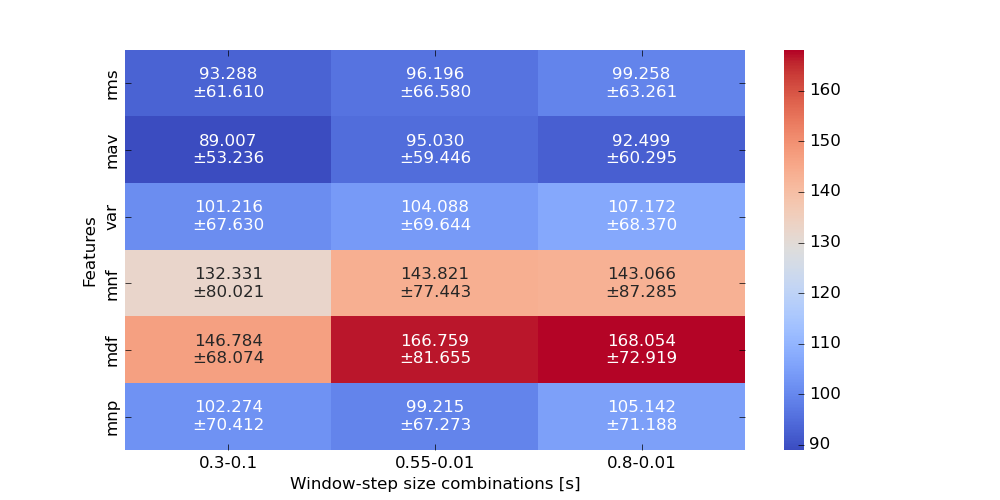}
        \caption{Loss, lower is better.}
    \end{subfigure}
    \hfill
    \begin{subfigure}{0.49\linewidth}
        \includegraphics[width=\linewidth]{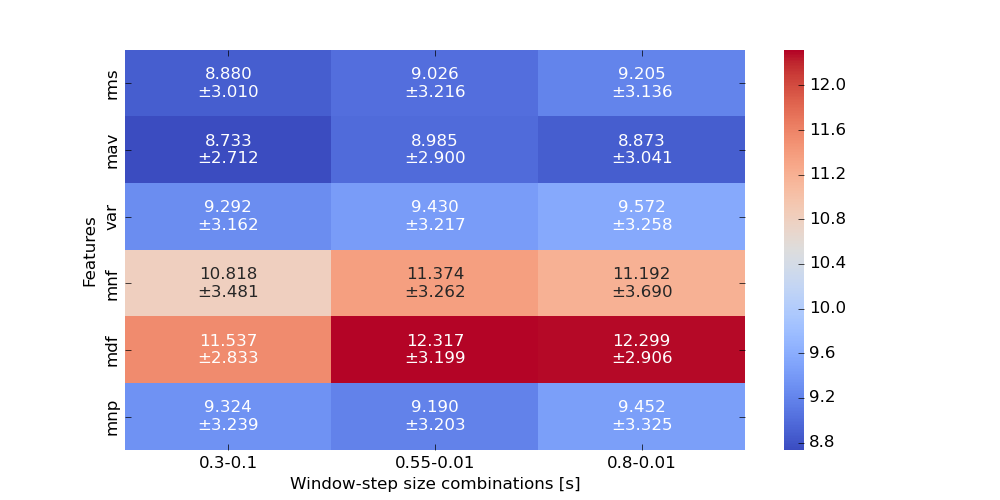}
        \caption{RMSE, lower is better.}
    \end{subfigure}
    
    \vspace{0.5em}

    \begin{subfigure}{0.49\linewidth}
        \includegraphics[width=\linewidth]{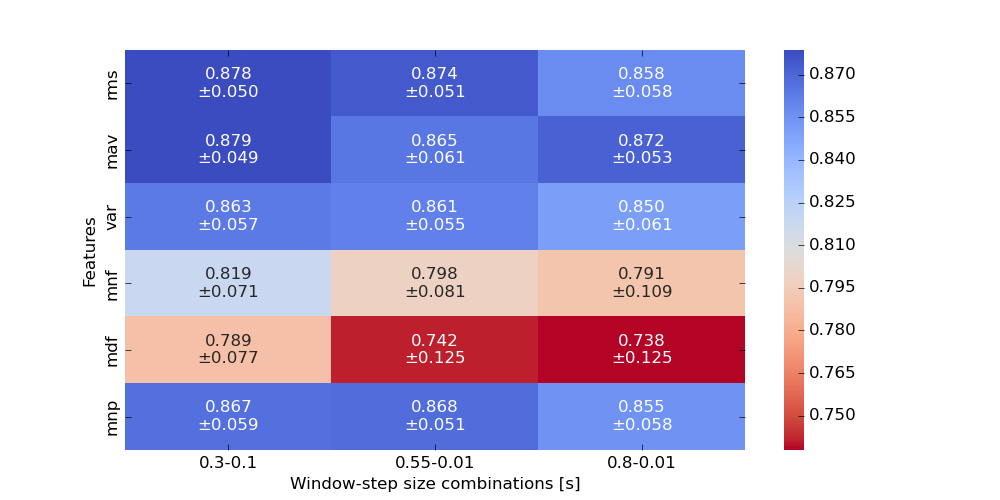}
        \caption{PCC, higher is better.}
    \end{subfigure}
    \hfill
    \begin{subfigure}{0.49\linewidth}
        \includegraphics[width=\linewidth]{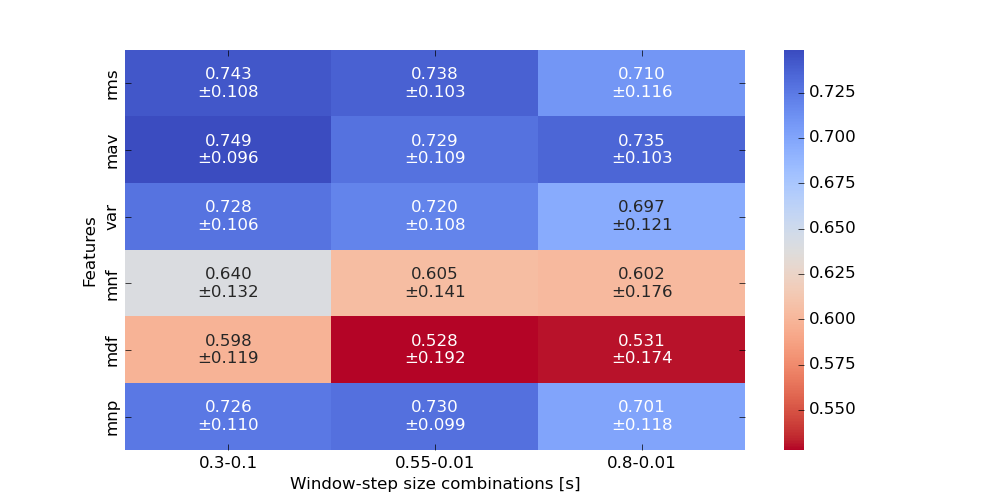}
        \caption{R2, higher is better.}
    \end{subfigure}

    \caption{Validation performance of window-step size and feature combinations. For all graphs darker blue indicates better performance.}
    \label{fig:heatmaps-feature-window-step-exp}
\end{figure}

Figure \ref{fig:heatmaps-feature-window-step-exp} shows the validation performance for each of the feature and window-step combinations. 
The time domain features tend to perform better than the frequency domain features.
When considering the standard error interval for each of the metrics, the time domain features tend to produce comparable results. 
The absolute ordering of features from best to worse are MAV, RMS, MNP, VAR, MDF, MNF; with the top 3 results obtained with
MAV with 0.3-0.1 window-step size, MAV with a 0.8-0.01 window-step combination, and RMS with a 0.3-0.1 window-step combination.
With the relatively large standard error, we do not consider the ordering statistically significant. 
The top three combinations outperforming MDF and MNF was statistically significant.


As with Figure \ref{fig:heatmaps-window-step-exp}, all four metrics agree on the relative ranking of feature and window-step combinations, for the same reasons discussed above. This agreement is notable here specifically because, unlike the window-step comparison (which varies only the amount and rate of data supplied to the same feature), the feature comparison varies the underlying signal representation itself; consistent agreement across metrics therefore gives stronger confidence that the observed performance differences reflect genuine differences in feature quality, rather than a metric-specific artefact.
The two best performing features, MAV and RMS, are recorded in Table \ref{tab:hpt-feature-final-stateless-rms-mav} as \textit{Phase 4 RMS} and \textit{Phase 4 MAV}, respectively.

\vspace{-10pt}

\subsubsection{Assessment of Subject Selection:}

Figure \ref{fig:series-exp-summary-feature-window-step} summarises the validation loss of individual subjects over all 18 experiments. 
In the figure, subjects 6 and 40 remain within their designated group (positions 5–6) throughout, but subjects 17 and 20 occasionally swap positions, 17 (normally 1–2) appearing at position 4, and 20 (normally 3–4) at position 2, for a few experiments. Subjects 39 and 9 remain within their groups throughout, with 9 never leaving position 1. This is more mixing than seen in the window-step experiment, indicating the six subjects are not wholly representative of the 35 when it comes to selecting features.

\vspace{-10pt}

\subsubsection{Latency}

\begin{figure}[h]
    \centering
    \includegraphics[width=0.6\linewidth]{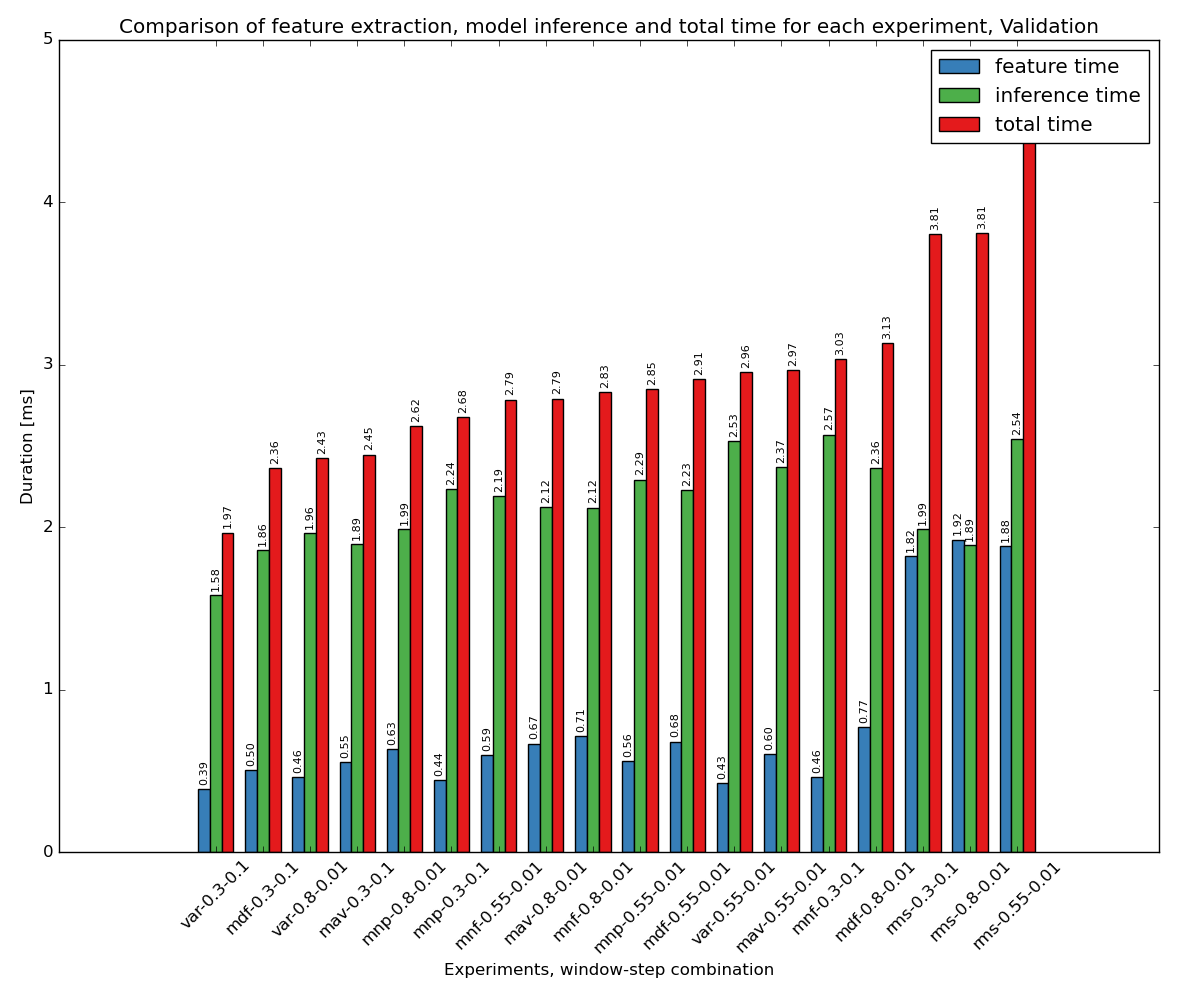}
    \caption{Extraction and inference time for each of the six features, at each of the three window-step combinations. Blue indicates feature extraction time, green indicates inference time, and red indicates the total time (the sum of the two).}
    \label{fig:series-exp-timings-bar-graph}
\end{figure}

Figure \ref{fig:series-exp-timings-bar-graph} shows the time it took for feature extraction, model inference and the total time between the two for each feature and window-step combination. The combination with the fastest feature extraction time is var-0.3-0.1, and the slowest feature extraction time is rms-0.8-0.01, confirming the results seen in Section \ref{sec:computational complexity}.

The quickest inference times are var-0.3-0.1 and mdf-0.3-0.1, despite differing computational complexity (O(N) vs O(N log N)) and domain.
The slowest are mnf-0.3-0.1 and rms-0.55-0.01 ms, an FD feature at the smallest window and a TD feature at a middle window size, suggesting inference is also independent of window size. This variation may instead be attributable to implementation-specific or hardware-level factors not captured by the complexity analysis, which are not investigated further here.

The longest time it took to extract a feature and inference is 4.4 ms (rms-0.55-0.01) while the shortest time is 2.0 ms (var-0.3-0.1). At most, the total time represents about 4\% of the total 125 ms controller delay constraint, effectively negligible. 

\subsection{Extended Evaluation}

The results are now reviewed with six additional subjects (numbers 3, 18, 23, 32, 33, and 37, selected from the same ranking as the original six), using HP tuning protocol B.
Table \ref{tab:hpt-feature-final-stateless-rms-mav} shows the average performance of the MAV (\textit{Phase 5 MAV}) and RMS (\textit{Phase 5 RMS}) features using a 300-100 window-step combination. 
Although MAV yielded better mean performance than RMS across all validation metrics, the differences were small relative to the standard error, indicating no meaningful distinction between the two features.
The models using both the MAV and RMS feature outperformed the baseline LSTM of Long et al.~\cite{long-2023-transfer-learning} across all metrics. 
These models also outperformed the LSTM-conv of Long et al.~\cite{long-2023-transfer-learning} (a more complex architecture newly developed in \cite{long-2023-transfer-learning}) in terms of both  $R^2$ and PCC, though not for RMSE. 
The comparison with the results of Long et al.~\cite{long-2023-transfer-learning} was statistically significant. 
However, as noted in Section \ref{sec:benchmark results}, direct comparison should be made with caution given differences in dataset composition and preprocessing.

Test set performance for both MAV and RMS was closely matched, within a standard error of each other, though both scored notably lower than on the validation set. 
This gap in performance is possibly due to the limited amount of data in the dataset, where the choices based on the validation set tailored the model to a very small and not wholly representative set of data.
Finally, we compare the test set performance of the two model sets (MAV and RMS) to their equivalent 12 subject baseline models from \textit{Phase 1 Baseline}. We isolate the 12 selected subjects from the 35 baseline models to obtain a comparable baseline. As can be seen in Table \ref{tab:hpt-feature-final-stateless-rms-mav}, both sets of models achieved significantly better performance. 
This performance improvement is seen over all four metrics, with clear improvement in $R^2$ and PCC and smaller improvement in loss and RMSE. 
These results suggest that optimizing the window size, step size, and feature type improves performance.


\begin{table}[t]
    \setlength{\tabcolsep}{6pt}
    \centering
    \caption{Summary of model test and validation performance.
    The number of subjects (S) and the HP tuning protocol (HHP) is noted along with the four performance metrics, loss, $R^2$, PCC, and RMSE. The best validation and test (for 12 subjects) performance is written in boldface, together with all values that lie within one standard error of the best value.}
    
    \resizebox{\textwidth}{!}{
    \small
    
    \begin{tabular}{ll | rrrr }
    \toprule
    
    Work & S - HPP & Loss & $R^2$ & PCC & RMSE  \\
    \midrule

    \multicolumn{6}{c}{\textbf{Test (35)}}\\
    \midrule
    
    Long et al. baseline & 35 - & - 
    & 0.632 $\pm 0.030$ & 0.749 $\pm 0.017$ & 9.666 $\pm 0.305$\\
    
    Long et al. LSTA-conv & 35 - & - 
    & 0.727 $\pm 0.041$ & 0.825 $\pm 0.015$ & 7.926 $\pm 0.325$ \\
    \midrule

    \multicolumn{6}{c}{\textbf{Test (12)}}\\
    \midrule

    Phase 1 Baseline & 12 - B 
    & 218.775 $\pm$ 91.774 & 0.389 $\pm$ 0.220 & 0.693 $\pm$ 0.112 & 13.868 $\pm$ 2.844 \\

    Phase 5 MAV & 12 - B 
    & \textbf{168.550} $\pm$ 70.535 & \textbf{0.544} $\pm$ 0.183 & \textbf{0.779} $\pm$ 0.080 & \textbf{12.181} $\pm$ 2.568 \\
    
    Phase 5 RMS & 12 - B  
    & \textbf{167.512} $\pm$ 72.967 & \textbf{0.540} $\pm$ 0.193 & \textbf{0.779} $\pm$ 0.085 & \textbf{12.112} $\pm$ 2.600 \\

    \midrule

    \multicolumn{6}{c}{\textbf{Validation}}\\
    \midrule
    
    Phase 1 Replication & 35 - 
    & 144.411 $\pm$ 51.992  & 0.537 $\pm$ 0.163 & 0.766 $\pm$ 0.097 & 11.459 $\pm$ 2.038 \\ 
    
    Phase 1 Baseline & 35 - B 
     & 94.840 $\pm$ 43.653 & 0.689 $\pm$ 0.117 & 0.848 $\pm$ 0.066 & 9.213 $\pm$ 2.005 \\

    Phase 2 Baseline & 6 - A
     & 122.855 $\pm$ 80.398 & 0.632 $\pm$ 0.142 & 0.817 $\pm$ 0.069 & 10.264 $\pm$ 3.389 \\
    \midrule

    Phase 4 MAV & 6 - A 
    & 89.007 $\pm$ 53.236 & 0.749 $\pm$ 0.096 & 0.879 $\pm$ 0.049 & 8.733 $\pm$ 2.712 \\
    Phase 4 RMS & 6 - A 
    & 93.288 $\pm$ 61.610 & 0.743 $\pm$ 0.108 & 0.878 $\pm$ 0.050 & 8.880 $\pm$ 3.010 \\

    Phase 5 MAV & 12 - B 
    & \textbf{78.893} $\pm$ 44.872 & \textbf{0.768} $\pm$ 0.089 & \textbf{0.890} $\pm$ 0.043 & \textbf{8.291} $\pm$ 2.320 \\
    Phase 5 RMS & 12 - B 
    & \textbf{84.324} $\pm$ 48.175 & \textbf{0.761} $\pm$ 0.097 & \textbf{0.888} $\pm$ 0.047 & \textbf{8.555} $\pm$ 2.377 \\
    
    \bottomrule
    \end{tabular}

    }
    
    \label{tab:hpt-feature-final-stateless-rms-mav}
\end{table}

\section{Conclusion}

This work systematically analysed different feature extraction options for an LSTM on the Ninapro DB2 dataset. 
Of the combinations evaluated, a window size of 300 ms, a step size of 100 ms, and both the MAV and RMS features yielded the best validation performance. 
On an unseen test set, these features obtained an average PCC of 0.779 ± 0.080 and 0.779 ± 0.085, respectively.

Overall, time domain features as well as MNP tended to produce comparable results, but MNF and MDF performed worse. 
%
%
Adjusted pre-processing and consistent HP tuning also demonstrated gains,
underscoring the importance of systematic tuning over modifying architectural complexity alone.
Final models outperformed both the baseline LSTM and the LSTM-conv of Long et al.~\cite{long-2023-transfer-learning}. 

A small, carefully selected subset of subjects was used to represent a much larger subject pool for analysis. Evaluation during window and step size selection indicated that the pool remained representative of trends,  though this representativeness weakened somewhat when evaluating different features.
Note that window size, step size, and feature choices were optimised specifically for this dataset and task; should the downstream task or data characteristics change, these parameters, and potentially the feature itself, should be recalibrated rather than assumed to transfer directly.
Feasibility testing showed that all window size, step size, and feature combinations could be deployed on an embedded system without exceeding the controller delay constraint of a smart prosthesis, with the best-performing combination contributing no more than a few milliseconds.


Future work could include more extensive HP tuning across all feature and window-step combinations, as well as investigating the use of combining features.
Further investigation into the effect of smaller window sizes would also be valuable, since the general pattern (where the smallest step size performed best for a given window size) was reversed at the smallest window size (300 ms), where the largest step size performed best instead.
This work could also be extended to multi-subject models, training on data from many subjects to create a generic finger-angle regressor, building on the direction taken by Long et al.~\cite{long-2023-transfer-learning}, whose LSTA-conv model similarly aimed to generalise from multiple subjects to an individual user.

Finally, these results have proven useful in a sister project, in which a prosthetic hand is being constructed with an actuation delay of approximately of 10 ms, the time the physical actuation system itself takes to move the prosthesis. Adding the total feature extraction and inference time determined in this work (approximately 3 ms) gives a combined delay of approximately 13 ms from signal to actuation, well within the controller delay constraint established earlier, allowing much flexibility in selecting window and step sizes.


\begin{credits}
\subsubsection{\ackname} 

This work is based on research supported in part by the National Research Foundation of South Africa (Grant Reference Number: RCDL240215206999) and SENTECH SOC Ltd.
Opinions expressed and conclusions arrived at are
those of the authors.

\subsubsection{\discintname}
The authors have no competing interests to declare that are
relevant to the content of this article. 
\end{credits}
%
%
%
\newpage
\bibliographystyle{splncs04}
\bibliography{sections/ref}

@misc{biewald-2020-wandb,
  author       = {Biewald, Lukas},
  title        = {Experiment Tracking with Weights and Biases},
  year         = {2020},
  howpublished = {\url{https://www.wandb.com/}},
  note         = {Software available from wandb.com}
}

@article{theunissen2025knowitrepo,
  title={KnowIt: Deep Time Series Modeling and Interpretation},
  author={Theunissen, MW and Rabe, R and Davel, MH},
  journal={arXiv preprint arXiv:2507.06009},
  year={2025}
}

@article{sid-el-moctar-2024-comprehensive-review-semg,
  author  = {Sid'El Moctar, Sidi Mohamed and Rida, Imad and Boudaoud, Sofiane},
  title   = {Comprehensive Review of Feature Extraction Techniques for sEMG Signal Classification: From Handcrafted Features to Deep Learning Approaches},
  journal = {IRBM},
  volume  = {45},
  number  = {6},
  pages   = {100866},
  year    = {2024},
  doi     = {10.1016/j.irbm.2024.100866}
}

@article{hochreiter-1997-long-short-term-memory,
  author  = {Hochreiter, Sepp and Schmidhuber, J{\"u}rgen},
  title   = {Long Short-Term Memory},
  journal = {Neural Computation},
  volume  = {9},
  number  = {8},
  pages   = {1735--1780},
  year    = {1997},
  doi     = {10.1162/neco.1997.9.8.1735}
}

@article{avian-2022-estimating-finger-joint,
  title={Estimating finger joint angles on surface EMG using manifold learning and long short-term memory with attention mechanism},
  author={Avian, Cries and Prakosa, Setya Widyawan and Faisal, Muhamad and Leu, Jenq-Shiou},
  journal={Biomedical Signal Processing and Control},
  volume={71},
  pages={103099},
  year={2022},
  publisher={Elsevier}
}

@article{farrell-2007-optimal-controller-delay,
  author    = {Farrell, Todd R. and Weir, Richard F.},
  title     = {The Optimal Controller Delay for Myoelectric Prostheses},
  journal   = {IEEE Transactions on Neural Systems and Rehabilitation Engineering},
  year      = {2007},
  volume    = {15},
  number    = {1},
  pages     = {111--118}
}

@article{gao-2021-multi-dof-hand,
  author    = {Gao, Z. and Tang, R. and Huang, Q. and He, J.},
  title     = {A multi-DOF prosthetic hand finger joint controller for wearable sEMG sensors by nonlinear autoregressive exogenous model},
  journal   = {Sensors},
  year      = {2021},
  volume    = {21},
  number    = {8},
  pages     = {2576}
}

@article{chen-2021-semg-large-scale-temporal-conv,
  author    = {Chen, C. and Guo, W. and Ma, C. and Yang, Y. and Wang, Z. and Lin, C.},
  title     = {sEMG-based continuous estimation of finger kinematics via large-scale temporal convolutional network},
  journal   = {Applied Sciences},
  year      = {2021},
  volume    = {11},
  number    = {10},
  pages     = {4678}
}

@book{criswell-2010-cram,
  author    = {Criswell, Eleanor},
  title     = {Cram's Introduction to Surface Electromyography},
  edition   = {2},
  publisher = {Jones \& Bartlett Publishers},
  address   = {Burlington, MA},
  year      = {2010}
}

@article{vujaklija-2016-new-prosthetic-dev,
  author  = {Vujaklija, Ivan and Farina, Dario and Aszmann, Oskar C.},
  title   = {New developments in prosthetic arm systems},
  journal = {Orthopedic Research and Reviews},
  pages   = {31--39},
  year    = {2016}
}

@article{catacora-2023-three-doublediff,
  author  = {Catacora, V. A. and Guerrero, F. N. and Spinelli, E. M.},
  title   = {Three-electrode double-differential biopotential amplifier for surface EMG measurements},
  journal = {IEEE Transactions on Instrumentation and Measurement},
  volume  = {72},
  pages   = {1--8},
  year    = {2023}
}

@article{guerrero-2015-analysis-doubleDiff,
  author  = {Guerrero, F. N. and Spinelli, E. M. and Haberman, M. A.},
  title   = {Analysis and simple circuit design of double differential EMG active electrode},
  journal = {IEEE Transactions on Biomedical Circuits and Systems},
  volume  = {10},
  number  = {3},
  pages   = {787--795},
  year    = {2015}
}

@article{bi-2019-review-emg,
  author  = {Bi, L. and Guan, C. and others},
  title   = {A review on EMG-based motor intention prediction of continuous human upper limb motion for human-robot collaboration},
  journal = {Biomedical Signal Processing and Control},
  volume  = {51},
  pages   = {113--127},
  year    = {2019}
}

@article{sawangjai-2021-eeganet-ocular,
  author  = {Sawangjai, P. and Trakulruangroj, M. and Boonnag, C. and Piriyajitakonkij, M. and Tripathy, R. K. and Sudhawiyangkul, T. and Wilaiprasitporn, T.},
  title   = {EEGANet: Removal of ocular artifacts from the EEG signal using generative adversarial networks},
  journal = {IEEE Journal of Biomedical and Health Informatics},
  volume  = {26},
  number  = {10},
  pages   = {4913--4924},
  year    = {2021}
}

@article{kundu-2018-hand-wheelchair,
  author  = {Kundu, A. S. and Mazumder, O. and Lenka, P. K. and Bhaumik, S.},
  title   = {Hand gesture recognition based omnidirectional wheelchair control using IMU and EMG sensors},
  journal = {Journal of Intelligent \& Robotic Systems},
  volume  = {91},
  pages   = {529--541},
  year    = {2018}
}

@article{ambrosini-2021-robotic-emg-training,
  author  = {Ambrosini, E. and Gasperini, G. and Zajc, J. and Immick, N. and Augsten, A. and Rossini, M. and Ballarati, R. and Russold, M. and Ferrante, S. and Ferrigno, G. and others},
  title   = {A robotic system with EMG-triggered functional electrical stimulation for restoring arm functions in stroke survivors},
  journal = {Neurorehabilitation and Neural Repair},
  volume  = {35},
  number  = {4},
  pages   = {334--345},
  year    = {2021}
}

@article{Ozdemir,
  author  = {Ozdemir, M. A. and Kisa, D. H. and Guren, O. and Akan, A.},
  title   = {Hand gesture classification using time--frequency images and transfer learning based on CNN},
  journal = {Biomedical Signal Processing and Control},
  volume  = {77},
  pages   = {103787},
  year    = {2022}
}

@article{Hudgins-features,
  author  = {Hudgins, B. and Parker, P. and Scott, R. N.},
  title   = {A new strategy for multifunction myoelectric control},
  journal = {IEEE Transactions on Biomedical Engineering},
  volume  = {40},
  number  = {1},
  pages   = {82--94},
  year    = {1993}
}

@inproceedings{li-2015-identification-finger-force,
  author    = {Li, X. and Fu, J. and Xiong, L. and Shi, Y. and Davoodi, R. and Li, Y.},
  title     = {Identification of finger force and motion from forearm surface electromyography},
  booktitle = {2015 IEEE International Conference on Multisensor Fusion and Integration for Intelligent Systems (MFI)},
  pages     = {316--321},
  publisher = {IEEE},
  year      = {2015}
}

@article{lee-2022-emg-stroke-training,
  author  = {Lee, K.},
  title   = {EMG-triggered pedaling training on muscle activation, gait, and motor function for stroke patients},
  journal = {Brain Sciences},
  volume  = {12},
  number  = {1},
  pages   = {76},
  year    = {2022}
}

@article{ngeo-2014-continuous-kinematics,
  author  = {Ngeo, J. G. and Tamei, T. and Shibata, T.},
  title   = {Continuous and simultaneous estimation of finger kinematics using inputs from an EMG-to-muscle activation model},
  journal = {Journal of Neuroengineering and Rehabilitation},
  volume  = {11},
  pages   = {1--14},
  year    = {2014}
}

@book{Goodfellow-et-al-2016-deep-learning,
  author    = {Goodfellow, Ian and Bengio, Yoshua and Courville, Aaron},
  title     = {Deep Learning},
  publisher = {MIT Press},
  year      = {2016}
}

@inproceedings{guo-2019-lstm-joint-angles,
  author    = {Guo, W. and Wang, C. and Lin, C. and Wang, C.},
  title     = {Long short term memory model based continuous estimation of human finger joint angles},
  booktitle = {2019 IEEE International Conference on Real-time Computing and Robotics (RCAR)},
  pages     = {288--292},
  publisher = {IEEE},
  year      = {2019}
}

@misc{tuningplaybookgithub,
  author       = {Godbole, V. and Dahl, G. E. and Gilmer, J. and Shallue, C. J. and Nado, Z.},
  title        = {Deep Learning Tuning Playbook},
  year         = {2023},
  note         = {Version 1}
}

@inproceedings{koch-2020-regression-rnn,
  author    = {Koch, P. and Dreier, M. and Larsen, A. and Parbs, T. J. and Maass, M. and Phan, H. and Mertins, A.},
  title     = {Regression of hand movements from sEMG data with recurrent neural networks},
  booktitle = {2020 42nd Annual International Conference of the IEEE Engineering in Medicine \& Biology Society (EMBC)},
  pages     = {3783--3787},
  publisher = {IEEE},
  year      = {2020}
}

@inproceedings{anam-2020-estimation,
  author    = {Anam, K. and Avian, C. and Swasono, D. I. and Muttaqin, A. Z. and Ismail, H.},
  title     = {Estimation of finger joint movement based on electromyography signal using long short-term memory},
  booktitle = {2020 International Conference on Computer Engineering, Network, and Intelligent Multimedia (CENIM)},
  pages     = {86--90},
  publisher = {IEEE},
  year      = {2020}
}

@inproceedings{ilyas-2022-evaluation-gated-recurrent,
  author    = {Ilyas, Z. and Anam, K. and Avian, C. and Muttaqin, A. and Ramadhan, M. E. and others},
  title     = {Evaluation of gated-recurrent unit for estimating finger-joint angle using surface electromyography signal},
  booktitle = {2022 9th International Conference on Electrical Engineering, Computer Science and Informatics (EECSI)},
  pages     = {25--28},
  publisher = {IEEE},
  year      = {2022}
}

@article{Putro-2024-Estimating-finger-joint,
  author  = {Putro, N. A. S. and Avian, C. and Prakosa, S. W. and Mahali, M. I. and Leu, J. S.},
  title   = {Estimating finger joint angles by surface EMG signal using feature extraction and transformer-based deep learning model},
  journal = {Biomedical Signal Processing and Control},
  volume  = {87},
  number  = {Part B},
  pages   = {105447},
  year    = {2024}
}

@article{geng-2022-cnn,
  author  = {Geng, Y. and Yu, Z. and Long, Y. and Qin, L. and Chen, Z. and Li, Y. and Guo, X. and Li, G.},
  title   = {A CNN-attention network for continuous estimation of finger kinematics from surface electromyography},
  journal = {IEEE Robotics and Automation Letters},
  volume  = {7},
  number  = {3},
  pages   = {6297--6304},
  year    = {2022}
}

@article{long-2023-transfer-learning,
  author  = {Long, Y. and Geng, Y. and Dai, C. and Li, G.},
  title   = {A transfer learning based cross-subject generic model for continuous estimation of finger joint angles from a new user},
  journal = {IEEE Journal of Biomedical and Health Informatics},
  volume  = {27},
  number  = {4},
  pages   = {1914--1925},
  year    = {2023}
}

@online{Ninapro-2025,
  author = {{Ninapro Team}},
  title  = {Ninapro: Non-Invasive Adaptive Prosthetics Database},
  year   = {2025},
  url    = {https://ninapro.hevs.ch/index.html},
  note   = {Accessed 5 June 2025}
}

@article{atzori-2014-electromyography-db2,
  author  = {Atzori, M. and Gijsberts, A. and Castellini, C. and Caputo, B. and Hager, A. G. M. and Elsig, S. and Giatsidis, G. and Bassetto, F. and M{\"u}ller, H.},
  title   = {Electromyography data for non-invasive naturally-controlled robotic hand prostheses},
  journal = {Scientific Data},
  volume  = {1},
  number  = {1},
  pages   = {1--13},
  year    = {2014}
}

@online{britannica-coefficient-of-determination,
  author = {{The Editors of Encyclopaedia Britannica}},
  title  = {Coefficient of Determination},
  year   = {2025},
  url    = {https://www.britannica.com/science/coefficient-of-determination},
  note   = {Accessed 4 July 2025}
}

\newpage

\appendix
\section{Equations}

\subsection{Performance metrics}
\label{app:metrics}
For a set of $N$ samples, let $y_k$ denote the target value and $\hat{y}_k$ the model prediction for sample $k$, with $\bar{y}$ and $\bar{\hat{y}}$ their respective means.

The coefficient of determination is
\begin{equation}
    R^2 = 1 - \frac{\sum_{k=1}^{N} (y_k - \hat{y}_k)^2}{\sum_{k=1}^{N} (y_k - \bar{y})^2},
    \label{eq:r2-score}
\end{equation}
the Pearson Correlation Coefficient is
\begin{equation}
    \text{PCC} = \frac{\sum_{k=1}^{N} (\hat{y}_k - \bar{\hat{y}})(y_k - \bar{y})}
    {\sqrt{\sum_{k=1}^{N} (\hat{y}_k - \bar{\hat{y}})^2}\,\sqrt{\sum_{k=1}^{N} (y_k - \bar{y})^2}},
    \label{eq:pearson}
\end{equation}
and the Root Mean Square Error is
\begin{equation}
    \text{RMSE} = \sqrt{\frac{1}{N} \sum_{k=1}^{N} (\hat{y}_k - y_k)^2}.
    \label{eq:rmse}
\end{equation}

\subsection{Features}
\label{app:features}
For the time-domain features, let $x_k$ ($k = 1,\dots,N$) be the sEMG samples in a window of $N$ samples, with mean $\bar{x}$:
\begin{align}
    \text{MAV} &= \frac{1}{N} \sum_{k=1}^{N} |x_k|, \label{eq:mav}\\
    \text{RMS} &= \sqrt{\frac{1}{N} \sum_{k=1}^{N} x_k^2}, \label{eq:rms-feature}\\
    \text{VAR} &= \frac{1}{N-1} \sum_{k=1}^{N} (x_k - \bar{x})^2. \label{eq:var}
\end{align}

For the frequency-domain features, let $P_k$ be the power spectrum value at frequency $f_k$, for $k = 1,\dots,M$ frequency bins (computed with the periodogram, Section~\ref{sec:computational complexity}):
\begin{align}
    \text{MNF} &= \frac{\sum_{k=1}^{M} f_k P_k}{\sum_{k=1}^{M} P_k}, \label{eq:mnf}\\
    \text{MDF} &= f_j \;\text{ such that }\; \sum_{k=1}^{j} P_k = \frac{1}{2} \sum_{k=1}^{M} P_k, \label{eq:mdf}\\
    \text{MNP} &= \frac{1}{M} \sum_{k=1}^{M} P_k. \label{eq:mnp}
\end{align}

\section{Training performance}


\begin{table}[htbp]
    \setlength{\tabcolsep}{6pt}
    \centering
    \caption{Model training performance in the same format as Table \ref{tab:hpt-feature-final-stateless-rms-mav}.
    }
    
    \resizebox{\textwidth}{!}{
    \small
    
    \begin{tabular}{ll | rrrr }
    \toprule
    
    Work & S - HPP & Loss & $R^2$ & PCC & RMSE  \\
    \midrule
    
    \multicolumn{6}{c}{\textbf{Training}}\\
    \midrule
    
    Phase 1 Replication & 35 - 
    & 11.526 $\pm$ 11.531 & 0.962 $\pm$ 0.043 & 0.982 $\pm$ 0.021 & 2.935 $\pm$ 1.530 \\
    
    Phase 1 Baseline & 35 - B 
    & 9.395 $\pm$ 7.535 & 0.973 $\pm$ 0.021 & 0.989 $\pm$ 0.008 & 2.735 $\pm$ 1.150 \\

    Phase 2 Baseline & 6 - A
    & 9.434 $\pm$ 6.471 & 0.975 $\pm$ 0.018 & 0.990 $\pm$ 0.008 & 2.764 $\pm$ 1.180 \\
    \midrule

    Phase 4 MAV & 6 - A 
    & 13.380 $\pm$ 3.319 & 0.960 $\pm$ 0.015 & 0.983 $\pm$ 0.007 & 3.548 $\pm$ 0.428 \\
    Phase 4 RMS & 6 - A 
    & 12.286 $\pm$ 6.529 & 0.967 $\pm$ 0.011 & 0.985 $\pm$ 0.005 & 3.339 $\pm$ 0.847 \\

    Phase 5 MAV & 12 - B 
    & 11.066 $\pm$ 5.198 & 0.968 $\pm$ 0.019 & 0.986 $\pm$ 0.009 & 3.137 $\pm$ 0.723 \\
    
    Phase 5 RMS & 12 - B 
    & 10.242 $\pm$ 5.945 & 0.973 $\pm$ 0.007 & 0.988 $\pm$ 0.007 & 2.973 $\pm$ 0.940 \\
    \bottomrule
    \end{tabular}

    }
    
    \label{tab:hpt-feature-final-stateless-rms-mav-train}
\end{table}

\section{Min-max normalisation}
\label{app:normalisation}

We were unable to find a rationale for the normalisation used by Long et al.~\cite{long-2023-transfer-learning} in related literature. 
To verify how the reported normalisation compares with a simpler approach, an HP sweep using Protocol~B was performed on the dataset of subject~20 (an average-performing subject, Section~\ref{sec:subject selection}).
The best model found for each method is reported in Table~\ref{tab:performance-best-long-ulaw}. The z-normalised model with $\mu$-law scaling removed outperformed all models on all validation metrics. As this is a single-subject comparison, it is indicative rather than conclusive; z-normalisation was nevertheless adopted instead of min-max normalisation, and $\mu$-law scaling was removed for all subsequent experiments.

\begin{table}[b!]
    \setlength{\tabcolsep}{6pt}
    \centering
    \caption{Performance when HP tuning the three pre-processing approaches: pre-processing method of Long et al. vs replacing min-max normalisation with z-norm vs also removing $\mu$-law scaling; result for subject 20, using protocol B.}
    \begin{tabular}{l| rrrr | cc}
    \toprule
    Experiment & Loss & R2 & PCC & RMSE & Inference& Epoch \\
     & & & & & time &  \\
    \midrule
    
    \multicolumn{7}{c}{\textbf{Validation}} \\    
    \midrule

    Tuned Long et al. & 116.280 & 0.543 & 0.744 & 10.229 & 0.003 & 94 \\
    + z-norm & 91.499 & 0.642 & 0.806 & 9.104 & 0.002 & 25 \\
    + $\mu$-law removed & \textbf{80.466} & \textbf{0.689} & \textbf{0.840} & \textbf{8.514} & 0.002 & 126 \\
    
    \midrule
    \multicolumn{7}{c}{\textbf{Training}} \\
    \midrule

    Tuned Long et al. & 51.548 & 0.793 & 0.914 & 6.840 & 0.017 & 94 \\ 
    + z-norm & 43.906 & 0.792 & 0.894 & 6.457 & 0.016 & 25 \\
    + $\mu$-law removed & \textbf{3.378} & \textbf{0.985} & \textbf{0.993} & \textbf{1.765} & 0.015 & 126 \\   

    \bottomrule
    \end{tabular}
    \label{tab:performance-best-long-ulaw}
\end{table}

\end{document}